\documentclass[11pt]{article}
\usepackage{ifpdf}
\usepackage{cite}
  \usepackage[pdftex]{graphicx}
  \graphicspath{{D/}}
  \DeclareGraphicsExtensions{.pdf,.jpeg,.png}

\usepackage{epstopdf}
\usepackage{caption}
\usepackage[cmex10]{amsmath}
\usepackage{amssymb}
\usepackage{algorithmic}
\usepackage{algorithm}
\usepackage{array}
\usepackage{multirow}
\usepackage{multicol}
\usepackage{mdwmath}
\usepackage{mdwtab}
\usepackage[tight,footnotesize]{subfigure}

\newtheorem{proposition}{Proposition}
\newtheorem{lemma}{Lemma}
\newtheorem{definition}{Definition}
\newtheorem{remark}{Remark}

\begin{document}

\title{Free Form based active contours for image segmentation and free space perception}

\author{Ouiddad~Labbani-I.$^{1,2}$, 
	Pauline~Merveilleux-O.$^{1}$ 
	and Olivier~Ruatta$^{2}$%
\thanks{*This work was partially supported by ANR agency under RDiscover project.}
\thanks{$^{1}$The authors are with Mod{\' e}lisation, Information et Syst{\`e}mes lab.,
        University of Picardie Jules Verne, Amiens, France
        {\tt\footnotesize \{pauline.merveilleux,ouiddad.labbani\}@u-picardie.fr}}%
\thanks{$^{2}$The authors are with XLim, UMR 7252 University of Limoges/CNRS, France. During this work, O. Labbani-Igbida was in CNRS research delegation with the XLim Institute.
{\tt\footnotesize \{ouiddad.labbani,olivier.ruatta\}@unilim.fr}%
       }
}
\maketitle
\begin{abstract}
In this paper we present a novel approach for representing and evolving deformable active contours.  The method combines piecewise regular B{\'e}zier models and curve evolution defined by local Free Form Deformation. The contour deformation is locally constrained which allows contour convergence with almost linear complexity while adapting to various shape settings and handling topology changes of the active contour. 

We demonstrate the effectiveness of the new active contour scheme for visual free space perception and segmentation using omnidirectional images acquired by a robot exploring unknown indoor and outdoor environments. Several experiments validate the approach with comparison to state-of-the art parametric and geometric active contours and provide fast and real-time robot free space segmentation and navigation.
\end{abstract}

\section{Introduction} \label{Sec.intro}

Free space perception, without any prior knowledge, is still a challenging issue in mobile robotics. Contributions to this challenge are numerous and typically used telemetric sensors, such as sonars or laser, to get an immediate perception of the objects depth and avoid collisions. Recent approaches use rather vision and mainly stereovision \cite{Knoeppel00, Rankin05, Agrawal06}, in which visual cues permit to recover this depth estimation. Only few contributions \cite{Murali08, Ohnishi06, Rochery06} consider monocular vision approaches. They however made strong assumptions on the camera motion and mostly required a metric or a topological knowledge of the environment to correctly perform the segmentation. Here, we investigate the use of active contours and monocular omnidirectional vision as a framework within which to realize real-time omnidirectional free space approximations in unknown environments.  We propose a new approach to contour parametrization and evolution based on Free Form Deformation.  Besides the benefit of offering a wide field of view, omnidirectional vision helps the initialization of contour propagation, that is a common problem in such approaches.

Contour extraction schemes based on deformable models are commonly and efficiently used in image segmentation.  They can be classified as either parametric \cite{Kass88} or geometric (variational) active contours \cite{Osher88,Caselles93} depending on the contour representation and its dynamic of propagation. The parametric active contours are explicit  active contour models whose deformation is constrained by minimization of a functional energy under a Lagrangian formulation.  In contrast, the variational methods consider implicit representation of the contour as the level sets of two dimensional distance functions which evolve according to an Eulerian formulation.  The seminal paper on active contours \cite{Kass88} considers parametric models.  This spawned many variations and extensions for energy formulations and contour dynamics \cite{Terzopoulos88, Cohen91, Williams92} in many imaging problems, including edge detection \cite{Davatzikos95}, shape modeling \cite{McIrnerney94,Terzopoulos88} and motion tracking \cite{Leymarie93,Terzopoulos93} to mention only few.  Whatever energy formulation \cite{Xu98, Sum07, Wu10}, the main concern is to ensure convergence of the contour to boundary concavities of the segmented object.  
Variational based contours\cite{Caselles93, Malladi95} reformulate the contour evolution in the context of PDE-driven surfaces and geometric flows. They overcome contour convergence in boundary concavities and allow to automatically handle topology changes of the contour using the level set framework. The approach proved to be efficient in a wide range of domains such as fluid dynamics or computer vision \cite{Bogdanova07, Chan02, Ying09} but with a prohibitive computational cost. The latter prevents these methods to be used in real time applications even with efficient numerical techniques as the Fast Marching methods \cite{Sethian99}.  Later, the authors \cite{Caselles97} proposed a geodesic approach that allows to connect classical active contours based on energy minimization with geometric active contours based on the theory of curve evolution. Close to this work is \cite{Precioso02, Bischoff04} which proposed piecewise regular contour models mixing spline functions and global minimization \cite{Precioso02}  or snaxels constrained motion \cite{Bischoff04}.

This work proposes a novel class of active contours that unifies piecewise regular contour models and curve evolution based on Free Form Deformation. It develops a new and fast technique for active contour parametrization and evolution. The active contour representation is given by parametric B{\'e}zier curve patches. These patches are referred to as Free-Form curves which suggest that they can be locally deformed with the flexibility and smoothness of B{\'e}zier curves of an arbitrary order.

The main contribution of the paper is to define the Free Form Deformation setup for active contour parametrization and propagation. The approach can adapt to various shape settings and handle topology changes of the active contour with almost a linear complexity. The linear complexity of our algorithms is upset only in case of topology change and is bounded in all cases by $\mathcal{O}(N\log N$) ($N:$ the number of the B{\'e}zier patches composing the active contour). A second contribution of the paper consists in using the Free Form based active contours framework for free space segmentation and safe navigation of a robot using catadioptric omnidirectional vision only. While the success of classical active contour models mostly depends on the initial contour locations with respect to the object to be segmented, the geometric properties of omnidirectional vision allows to encapsulate the initial contour without any prior knowledge on the shape or extent of the free space. A comparative study with recent work based on parametric and geometric models, validates the approach for real time free space segmentation and navigation in unknown and large environments.

The paper is structured as follows. Section~\ref{Sec.FF} recalls some basic notions about B{\'e}zier curves interpolation and more advanced concepts on Free Form deformation that will be used in active contour evolution. The latter is introduced in Section~\ref{Sec.FFAC} as well as some geometric operations allowing contour convergence and fitting to complex shapes and topology changes. Section~\ref{Sec.results} gives results, first in case of simplified synthetic images to illustrate the proposed method, and then in real world robot experiments that is a more advanced and challenging setting for the method assessment. It is evaluated on images acquired by a mobile robot exploring unknown indoor and outdoor environments and compared to state-of-the-art geometric and parametric models. Comparative work \cite{Merveilleux11a, Merveilleux11b} is used as a baseline for performance evaluation both in terms of processing time and contour convergence. Finally, Section~\ref{Sec.conclu} concludes this study and draws some future work.

\section{B{\'e}zier curves and free forms} \label{Sec.FF}

This section provides the material and basic tools needed for the free form modeling and deformation of active contours. The latter is composed of several linked patches, each of them is described by a continuous B{\'e}zier curve of order $d$, namely, \emph{Free Form}. 

\subsection{Curve parametrization}

A B{\'e}zier curve is defined by its order, $d$, and a set of \mbox{$d+1$} control points $P_0,\ldots, P_d$, where points are represented as vectors in $\mathbb{R}^2$. The first and the last control points are always the end points of the curve (that is the endpoint interpolation property); however, the intermediate control points (if any) generally do not lie on the curve. 

Let denote by $B([P_0, \ldots, P_d], t)$ the B{\'e}zier curve of order $d$ and consider a recursive definition as given, for $ t \in [0, 1]$, by
\begin{equation}
\left\{\!\!
\begin{array}{l}
{\displaystyle B([P_0], t) = P_0}\\
\!\!\!\begin{array}{l} {\displaystyle B([P_0, \ldots,P_d], t) = (1 - t)\ B([P_0, \ldots, P_{d - 1}], t) } \\ 
\hspace{3.5cm} {\displaystyle +\; t\, B([P_1, \ldots, P_d],t) } \end{array}
\end{array}
\right.
\label{Eq.Bezier}
\end{equation}
which expresses a B{\'e}zier curve of order $d$ as a linear interpolation between two B{\'e}zier curves of order $d-1$. 

Let also introduce the Bernstein polynomials of degree $d$ defined by 
\begin{equation}
{\displaystyle b_{d, i} (t) = \binom{d}{i} t^i (1 - t)^{d - i},\ i \in \{0, \ldots, d\} }
\label{Eq.Bern}
\end{equation}
and known as Bernstein basis polynomials of degree $d$ (where $\binom{d}{i}$ are binomial coefficients). It is easy to see that 
\begin{equation}
{\displaystyle B ([P_0,\ldots, P_d], t) = \sum^d_{i = 0} b_{d, i} (t) P_i }
\label{Eq.BezierB}
\end{equation}
with ${\displaystyle P_i,i=0,\ldots,d}$, the B{\'e}zier control points. 

\begin{remark}\label{Rq.convex}
Eq.~\eqref{Eq.BezierB} means that each point of a B{\'e}zier curve, ${\displaystyle B ([P_0,\ldots,P_d], t)}$, is a weighted average of the control points $P_i$, hence it lies inside the convex hull of those points.
\end{remark}

\begin{remark}\label{Rq.bases}
Seeing Eq.~\eqref{Eq.BezierB} and that the Bernstein basis polynomials of degree $d$ form a basis for the vector space of polynomials of degree $d$, 
so, every curve admitting a polynomial parametrization of degree $d$ can also be represented as a B{\'e}zier curve.
\end{remark}

B{\'e}zier curves could be evaluated using the \mbox{\emph{De Casteljau}} algorithm \cite{Farin00}, which is known to be the most robust and numerically stable method for evaluating polynomials. The corresponding scheme used the recursive definition \eqref{Eq.Bezier} and is given in Algorithm \ref{Algo.casteljau}.

\begin{algorithm}
\caption{De Casteljau algorithm : $Eval(B(t))$, \cite{Farin00}.}
\label{Algo.casteljau}
\begin{footnotesize}
\begin{algorithmic}
\REQUIRE $\left[ P_0,\!\dotsc\!P_n\right]$ the list of the points of the control polygon of the B{\'e}zier curve $B ([P_0,\ldots, P_d], t)$ and $t \in \left[0,1\right]$
\ENSURE Coordinates of a curve point in $\mathbb{R}^2$
\smallskip
\IF{$n=0$} \STATE \textbf{return} $P_0$
\ELSE
\STATE \textbf{return} $(1-t)*Eval(\left[P_0,\!\dotsc\!P_{n-1}\right],t)+t*Eval(\left[P_1,\!\dotsc\!P_n\right],t)$
\ENDIF
\end{algorithmic}
\end{footnotesize}
\end{algorithm}


\subsection{Curve interpolation}

Given a (uniform) sampling $M_i=B([P_0,\ldots, P_d],t_i)$ of a B{\'e}zier curve\footnote{We know that such a curve exists from Remark \ref{Rq.bases}.},  with $t_0\!=\!0\!<\!t_1\!<\!\dotsc\!<\!t_{d - 1}\!<\!t_d\!=\!1$ a subdivision of $[0,1]$, one can try to find the control points $P_i$ such that,
\begin{equation}
{\displaystyle 
\left(\!\!\begin{array}{c} M_0\\ \vdots\\ M_d \end{array}\!\!\right) = 
V\cdot \left(\!\!\begin{array}{c} P_0\\ \vdots\\ P_d \end{array}\!\!\right)
}\label{Eq.linear}
\end{equation} 
with $V$ the Bernstein basis polynomials evaluated at the $t_i$ samples:
\begin{equation}
{\displaystyle 
V = \left(\!\!\begin{array}{cccc}
     b_{d,0} (0) & b_{d,1} (0) & \cdots & b_{d,d} (0)\\
     b_{d,0} (t_1) & b_{d,1} (t_1) & \cdots & b_{d,d} (t_1)\\
     \vdots & \vdots &  & \vdots\\
     b_{d,0} (t_{d-1}) & b_{d,1} (t_{d-1}) & \cdots & b_{d,d} (t_{d-1})\\
     b_{d,0} (1) & b_{d,1} (1) & \cdots & b_{d,d} (1)
   \end{array}\!\!\right)
}\label{Eq.V}
\end{equation}
This gives the interpolation formula for the B{\'e}zier points. The inverse $V^{- 1}$ defines a linear mapping associating to $d+1$ distinct points, a B{\'e}zier curve of degree $d$ going through those points.

\begin{proposition}[Curve interpolation]\label{Prop.AC} 
Let $M_0,\ldots,M_d\in\mathbb{R}^2$ be distinct points and $t_0\!=\!0\!<\!t_1\!<\!\dotsc\!<\!t_d\!=\!1$ a subdivision of $[0, 1]$ and let $P_0,\ldots,P_d\in \mathbb{R}^2$ be given by:
\begin{equation}
\left(\!\!\begin{array}{c} P_0\\ \vdots\\ P_d \end{array}\!\!\right) = V^{- 1}\cdot 
\left(\!\!\begin{array}{c} M_0\\ \vdots\\ M_d \end{array}\!\!\right)
\label{Eq.AC}
\end{equation}
Then, if $\Gamma(t) = B([P_0,\ldots,P_d], t)$, we have $\Gamma (t_i) = M_i$, $\forall i\in\{0, \ldots, d\}$. 
\end{proposition}

Remark that for a given fixed subdivision $\{t_i\}_i$ of the interval $[0,1]$, the computation of the inverse mapping $V^{-1}$ is to be performed only once (and offline) in the initialization.  
So, the computational cost of the interpolation at degree $d$ is given by the following lemma:
\begin{lemma} \label{cost1}
	Assuming a $\{t_i\}_i$ fixed partition, the cost of interpolation by a B\'ezier curve of degree $d$ is the cost of multiplications of a $(d\!+\!1)\!\times\!(d\!+\!1)$ square matrix by two $(d\!+\!1)$-dimensional vectors. The related cost is in $\mathcal{O}(d^2)$ and is constant for a fixed $d$.
\end{lemma}

\subsection{Curve deformation}

\begin{figure}[h]
\centering
\subfigure{\includegraphics[width=.6\linewidth]{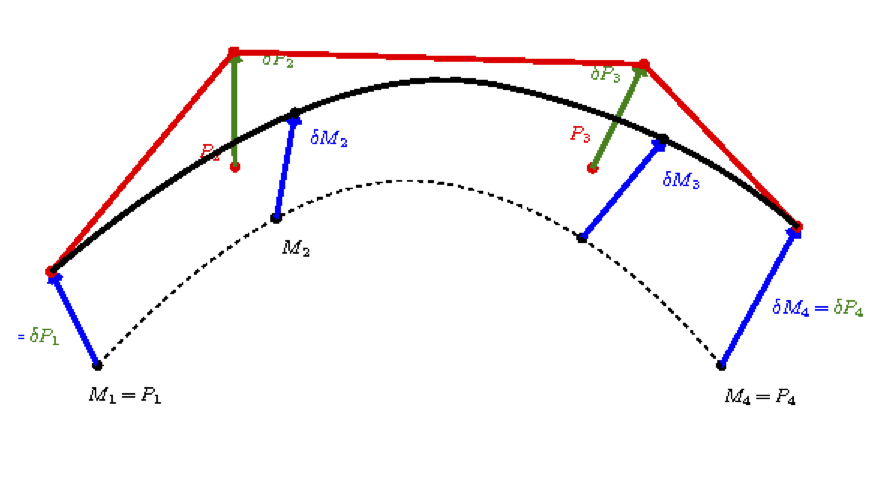}} \\ 
\subfigure{\includegraphics[width=.6\linewidth]{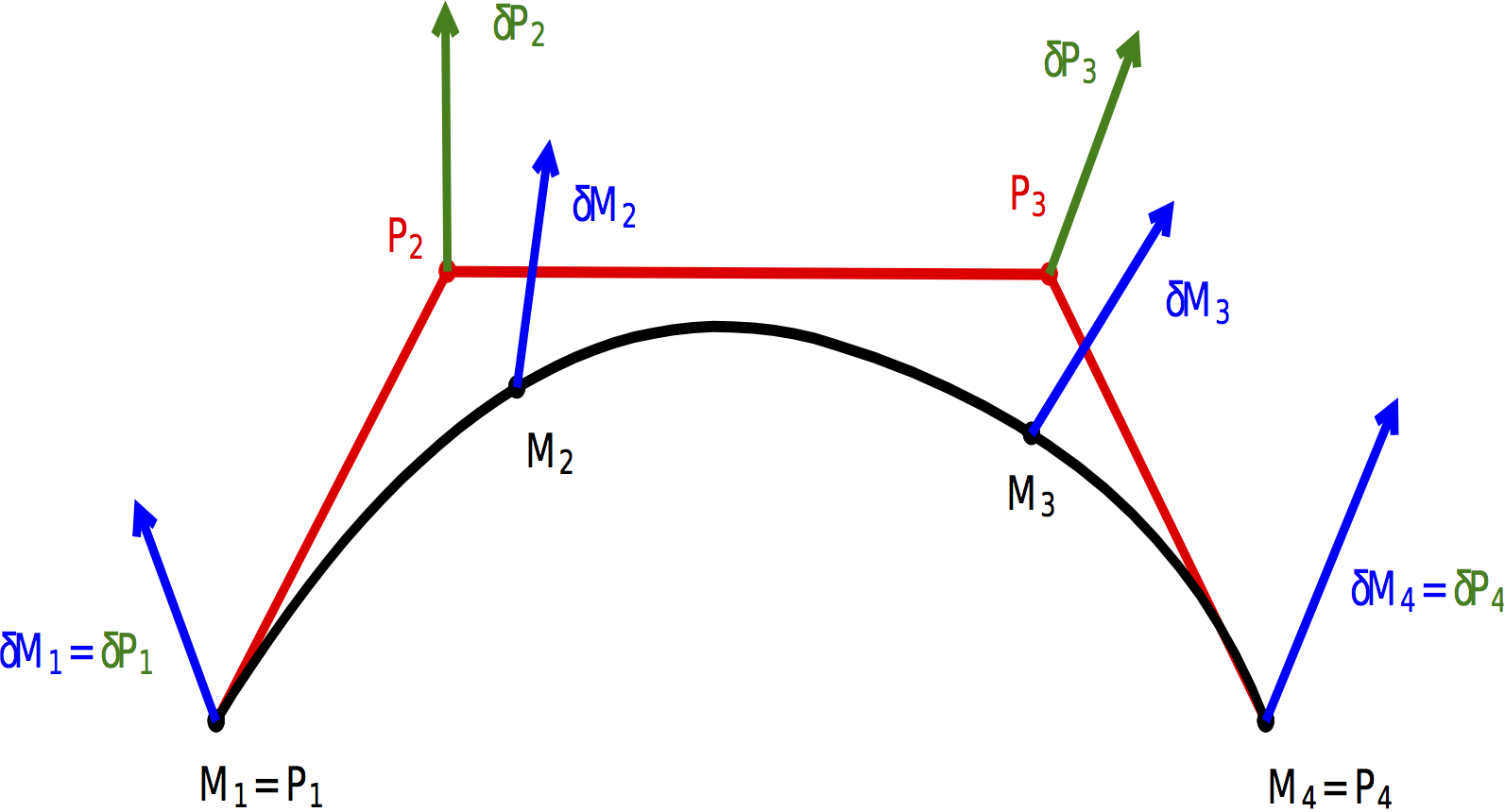}}
\caption{B{\'e}zier curve deformation. The undeformed curve appears at the \emph{bottom}, and the \emph{top} shows the lifting deformation on the control points.}
\label{Fig.deform}
\end{figure}

Let now consider a local deformation of the B{\'e}zier curve points (Fig.~\ref{Fig.deform}). The new curve is defined to be the resulting deformed control points correspondingly to the following proposition.

\begin{proposition}[Curve deformation]\label{Prop.deformpatch}
Let 
${\displaystyle (\delta M_0,\ldots,\delta M_d)}$ be a perturbation of the points ${\displaystyle \Gamma(t_0)\!=\!M_0,\!\ldots,\!\Gamma(t_d)\!=\!M_d}$, then the control points of the B{\'e}zier curve going through ${\displaystyle M_i+\delta M_i}$ at $t_i$ are given by
\begin{small}
\begin{equation}
\left(\!\!\begin{array}{c} P_0\\ \vdots\\ P_d \end{array}\!\!\right) + \left(\!\!\begin{array}{c} \delta P_0\\ \vdots\\ \delta P_d \end{array}\!\!\right)  \hbox{ with } \left(\!\!\begin{array}{c} \delta P_0\\ \vdots\\ \delta P_d \end{array}\!\!\right)  = V^{- 1}.\left(\!\!\begin{array}{c} \delta M_0\\ \vdots\\ \delta M_d \end{array}\!\!\right)
\label{Eq.deltaP}
\end{equation}
\end{small}
\end{proposition}

The same remark as for curve interpolation holds, the matrix $V^{-1}$ can be computed only once. The following lemma is a direct consequence of the remark:
\begin{lemma} \label{cost2}
	The cost of deformation of the control polygon of a B\'ezier curve of degree $d$ can be computed with the cost of multiplications of a $(d\!+\!1)\!\times\!(d\!+\!1)$ square matrix by two $(d\!+\!1)$-dimensional vectors and so has cost $\mathcal{I}(d)$.
\end{lemma}






\section{Free Forms for active contour deformation} \label{Sec.FFAC}




B\'ezier curves have several advantages but carry the rigidity of polynomial curves. In order to increase geometric possibilities, we can connect together B\'ezier curves. This leads to the notion of Free Form (FF for short). In this section, we define free forms and introduce the geometric operations needed on this type of structures in order to apply them to deformable contours. 

\subsection{FF-based active contour definition}

An active contour will be expressed as a set of $N$ linked continuous B{\'e}zier curves of the same order $d$ where the $N^{th}$ curve is connected to the first one (a closed form). 
Each B{\'e}zier curve defines a patch of the active contour; the latter being noted $\Gamma(t)\equiv\{\Gamma_j(t)\}$ with $j\!=\!1,\!\dotsc\!N$ a patch number, and called a free form active contour. 

\begin{definition}[Free Form]\label{Def.FF}
For $N, d$ integers, consider \mbox{$\mathcal{P}=\left[P_0,\!\dotsc\!P_{d-1}, P_d,\!\dotsc\!P_{2 d},\!\dotsc\!P_{Nd-1} \right]$} some points in $\mathbb{R}^2$. 
\mbox{We denote ${\displaystyle \Gamma_i(t) = B(\left[P_{(i-1)d},\!\dotsc\!P_{id}\right],t)}$ for $i=1,\!\dotsc\!N\!\!-\!\!1$}, and 
${\displaystyle \Gamma_{N}(t)=B(\left[ P_{(N-1) d},\!\dotsc\!P_{N d-1}, P_0\right],t)}$.\\
Let: 
\begin{equation*}
\Gamma_{d,N} : \left\{ \begin{array}{ccc} [0,1] & \longrightarrow & \mathbb{R}^2 \\ t & \longmapsto & \Gamma_i(Nt-i+1) \mbox{ if } t \in [\frac{i-1}{N},\frac{i}{N}] \end{array} \right.
\end{equation*}
Then $\Gamma_{d,N}$ is the free form associated to the list of points $\mathcal{P}$, it is a closed and continuous curve (since $\Gamma_i(1) = \Gamma_{i+1}(0)=P_{id}$ for all $i \in {1,\!\dotsc\!N-1}$ and $\Gamma_{N}(1)=\Gamma_{1}(0)=P_0$). It is a curve obtained by connecting together $N$ B\'ezier curves (each such curve is called a patch) of degree $d$. 
\end{definition}
Note that $t$ is the parameter that controls the distance ratio along interpolation: as it varies between $0$ and $1$, the entire curve could be generated.

From definition \eqref{Eq.Bezier}, we get an efficient way to compute and evaluate the parametrization of a free form using the De Casteljau algorithm for each patch of the active contour.  The regularity of B{\'e}zier curves brings an intrinsic smoothness to the active contour (continuous and differentiable everywhere but to a finite set of points).   One can add constraints so as to maintain higher cross-boundary derivative continuity. The latter is not considered here since we did not use them in our experiments but should be explored in future works.


\subsection{FF-based active contour evolution}


Classical parametric formulation of active contour models defines internal and external forces to evolve the contour. The external forces derive from image features whereas internal forces ensure smoothness and regularity constraints on the contour propagation. In the proposed work, the formulation is simplified by using Free Form deformation models. Indeed, it is not necessary to include regularization constraints as they are intrinsically derived from B{\'e}zier curves. 

The contour evolution is induced by local deformations of some selected points on the free form curves composing the patches of the active contour. 
A local deformation $\delta M_i$ at a selected contour point $M_i$ is only derived from a pressure force and a gradient edge map of the image.  

The edge map is defined such that it has large values only in the immediate neighborhood of edges, almost zero values in homogeneous regions and that the gradient vectors are pointing toward and are normal to the edges. It could be obtained using Canny edge detector \cite{Canny86}. This defines a computational diffusion process (Eq.~\eqref{Eq.edge}) in the image to attract free forms to regions boundaries.  The active free form contour is then attracted to edges or terminations. 
\begin{equation}
F_{diff} = \frac{1}{1+ F_{edge}}
\hbox{ with  }
F_{edge} = \vert\nabla G_{\sigma}\star I\vert^p; p\geq 1
\label{Eq.edge}
\end{equation}
where $G_{\sigma}\star I$ is a gaussian smoothing of image $I$ and $p\in\mathbb{N}$.

A balloon force (similarly to \cite{Cohen91}) is defined a pressure force and is used to move the contour in the normal direction to the B{\'e}zier curve defining each contour patch. Its effect is defined only in the homogenous domains of the gradient edge map. In our method, we take benefit of the closed form expression of B{\'e}zier curves to compute normal vectors at the selected contour points. This is achieved by computing (only once) the second derivative of the interpolation equation \eqref{Eq.BezierB}; that is easy since the control points are constants which reduces to the computation of the derivatives of Bernstein's polynomials.  Note that with some simple algebraic manipulations, the control points of B{\'e}zier derivative curves can be obtained immediately and can be computed at the same time with B{\'e}zier curves using de Casteljau algorithm.


Each point $M_{i,j}$ ($i=0,..,d$) of a contour patch $\Gamma_j$ is thus moved to a new position following the normal direction to the contour with a gradient function velocity. Control points $[P_{(j-1)d}, .., P_{jd}]$ of each patch $\Gamma_j$ at the new position are then obtained from equation \eqref{Eq.deltaP}, which in turn, define the entire B{\'e}zier curve parametrizing the new position of each active contour patch. The process is repeated until convergence (see free form algorithms below).  Note that the evolution of the active contour depends only on the evolution of these points,  whereas its regularity is maintained during propagation by the smoothness properties of B{\'e}zier curves mapping.

\subsection{Geometric refinement}

The initially defined active contour is composed of $N$ patches each of them of degree $d$. The contour expands to meet free space boundaries, but depending on the patches resolution, it could prevent from suitably converging in concave or convex areas. Considering that no model nor knowledge is given on the shape or extent of the free space, we make the number of patches dynamically evolve with active contour deformation. This is locally achieved by constraining the distance between adjacent points of the patches and inserting new patches where it is needed. We consider here the case of contour expansion\footnote{To operate pruning in case of contour shrinking, one could replace two adjacent patches by one patch formed by removing one point on two of the control points of the two patches}.

\begin{figure}[htbp]
\begin{center}
\includegraphics[width=8cm]{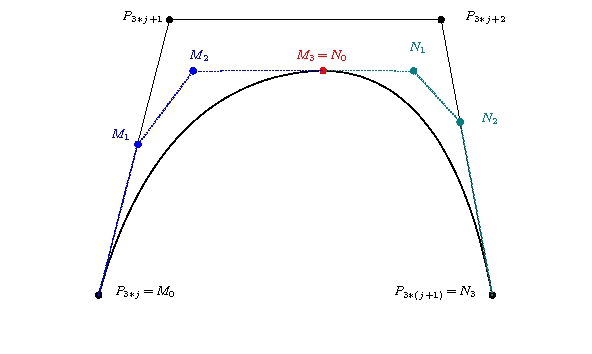}
\caption{Split and insert procedure. The distance between two adjacent control points (the \emph{(Red)} polygon vertices) exceeds a defined threshold $\varepsilon_d$. A split step is operated to generate two new patches (whose control polygons are in \emph{(Blue)} and \emph{(Green)}) from the initial B{\'e}zier curve \emph{(Black)}.}
\label{Fig.split}
\end{center}
\end{figure}

The insertion test is based on the maximal distance separating control points of the same patch: if it is above a fixed distance threshold, the patch is split into two different patches of the same degree $d$ as the initial one (see Fig.~\ref{Fig.split} which illustrates the cubic patches case). The latter are generated by uniformly resampling the initial B{\'e}zier curve in $2 d +1$ points $M_0,\ldots,M_{2 d}$ using De Casteljau algorithm, then interpolating each set of points $\{M_0,\ldots,M_{d}\}$ and $\{M_{d},\ldots,M_{2 d}\}$ to compute the respective control points of the new patches.

When an active is deformed, the distance test allows to determine if we need to split the patch in two. The test and the split are depicted in Algorithm~\ref{algo.split}. This leads the active contour to globally adapt to large or complex shapes of the free space boundaries, while maintaining a linear complexity of the algorithm on the total number of the active contour patches.


\begin{algorithm}
\caption{Split procedure.}
\label{algo.split}
\begin{footnotesize}
\begin{algorithmic}
\REQUIRE FF-based active contour $\Gamma_{d,N}$ (represented by the list of control points) and a threshold $\epsilon$.
\ENSURE New FF-based active contour (represented by the new list of control points)
\smallskip
\STATE $NCP \leftarrow []$
\STATE $nnp \leftarrow 0$
\FOR{$j \in \left\{1,\cdots,N\right\}$ }
\STATE $CP \leftarrow [P_{(j-1)*d},\cdots,P_{(j-1)*d+d}]$
\IF{$\max d(P_{k},P_{k'\neq k})  >  \epsilon$}
\STATE Compute $[M_0,\cdots,M_d]$ the control polygon of $\Lambda(t)$ by interpolation such that $\Lambda(\frac{i}{d})=B(CP,\frac{i}{2 d})$ for all $i \in \left\{ 0,\cdots,d \right\}$
\STATE Compute $[N_0,\cdots,N_d]$ the control polygon of $\Theta(t)$ by interpolation such that $\Theta(\frac{i}{d})=B(CP,\frac{i+d}{2 d})$ for all $i \in \left\{ 0,\cdots,d \right\}$
\STATE $CP \leftarrow [M_0,\cdots,M_d,N_0,\cdots,N_d]$
\STATE $nnp \leftarrow nnp + 1$
\ENDIF
\STATE Concatenate $NCP$ and $CP$
\ENDFOR
\STATE $N \leftarrow N + nnp$
\STATE \textbf{return}  $NCP$ //the list of the new control points  
\end{algorithmic}
\end{footnotesize}
\end{algorithm}


The following proposition bounds the cost of the split procedure:
\begin{lemma} \label{cost3}
	For each patch that needs to be split, the algorithm requires $2 d + 1$ evaluations and two interpolations, so the cost of the split procedure is in $\mathcal{O}(d^2)$ since the more expensive steps are given by the $2 d +1$ evaluations and the two interpolation steps. 
\end{lemma}

\subsection{Topology changes}

It is well known that geometric optimization does not deal with topology changes of the constrained forms. We here use geometric curve overlapping to manage topology changes of the free form. This is convenient to detect multiple isolated objects in the free space enclosing the robot.

In order to detect auto-intersection of the free form, we use a geometric filter as often in algorithmic geometry. Very often, to compute a predicate, the computation of the complete predicate is computationally very expensive even if it is very often false. This is the case when one as to test if some determinant vanish.  A filter is a computationally simpler predicate allowing to avoid to compute almost every time the expensive one avoiding to test in case of an obvious negative answer. One has to do the complex computation only when the result of the simpler one is positive. It allows to avoid a lot of computation. The efficiency of this approach depends on the cost of the simpler algorithm (the filter) and its selectivity (the probability or the rate of false positive answers). For a systematic description of this approach in algorithmic geometry see \cite{FP11}. 

We first describe the algorithms use for the geometric filter and then we focus on the flip algorithm. The global idea is to exclude a huge number of uninteresting tests using bounding boxes. In order to test if two patches can not present an intersection, we consider the bounding boxes of each patch. Let $\Gamma_i$ and $\Gamma_j$ be two distinct (and non successive) patches of active contour $\Gamma_{d,N}$, defined respectively by the sets of B{\'e}zier control points $\left\{P_{d i-d},P_{d i-d+1}, \cdots,P_{d i-1},P_{d i}\right\}$ and $\left\{P_{d j-d},P_{d j-d+1},\cdots,P_{d j-1},P_{d j}\right\}$. Trivially, each set of control points $\{P_{k}\}$ is contained in a rectangular bounding box which vertices are $(x_{k,min},y_{k,min})$, $(x_{k,max},y_{k,min})$, $(x_{k,max},y_{k,max})$ and $(x_{k,min},y_{k,max})$. Thus, two patches intersect only if their two corresponding bounding boxes intersect. This is the base of the {\bf bounding boxes filter}. 

If the bounding boxes intersect, it is possible that the control polygons do not intersect. Here, we do not test directly intersection of the curves, but the one of the control polygon. To test intersection of control polygons we use a simplified sweeping algorithm for line segments arrangement as in \cite{EG89}. For the sake of simplicity, we will make the assumption that the patches intersect properly, it is to say that the patches as two intersection points, each intersection being the proper intersection of two segments of each control polygon. 
The data structure of the full curve is important in order to achieve a good complexity. Here, we consider the curve $\Gamma_{d,N}$ has the list of the control polygons of its patches $[\Gamma_1,\ldots,\Gamma_N]$. Together with this list, we maintain a list $B = [B_1,\ldots,B_N]$ where $B_i = [x_{i,min},y_{i,min},x_{i,max},y_{i,max}]$ is the bounding box of $\Gamma_i$. The list $\Gamma_{d,N}$ is sorted using a lexicographical order on the bounding boxes. 
Let $B_1 =[x_1,y_1,x_2,y_2]$ and $B_2 = [x_3,y_3,x_4,y_4]$ be two boxes, we define $BoxInter$ as the function taking $B_1$ and $B_2$ as an input and returning $1$ if the boxes has an intersection and $0$ else. It a very simple algorithm since it return $0$  if one of the following condition is satisfied:
\begin{itemize}
\item[$\bullet$] $x_2 < x_3$, 
\item[$\bullet$] $x_2 = x_3$ and ($y_2 < y_3$ or $y_4 < y_1$),
\item[$\bullet$] $x_4 < x_1$,
\item[$\bullet$] $x_4 = x_1$ and ($y_4 < y_1$ or $y_2 < y_3$).
\end{itemize}
Else the function return $1$. 
If the enclosing boxes have a non-empty intersection, it does not mean that there is mandatory an intersection of the control polygon, neither of the curves. 
Here, we will describe algorithms for the case of patches of degree 3. There is two main motivations for that. Firstly, degree 3 is rich enough from the geometric point of view. Secondly, the algorithm for degree 3 patches is easy to understand and to describe while it is hard to give a completely general algorithm for the general case. So, we give here the complete algorithm for the cubic case that the reader can easily extend to the general case. The flip algorithm presented here consists in, starting from two control polygons with a non empty intersection, connect the extremities of one patch to the ones of the other by line segments respecting orientation given by the parametrization (it is insure by the fact that the two new control polygons do not intersect). 

\begin{remark}
If two patches met, we need to check the orientation of the patches are compatible. If two patches have an intersection and if the orientations are not compatible, it is impossible to change the connection without creating a new intersection. The only case where such a case can occur is when we have to merge two connected components of the curve. In order to make the connection possible, we have to reverse one of the parametrization. 
\end{remark}

\begin{figure}[h]
\includegraphics[width=5cm]{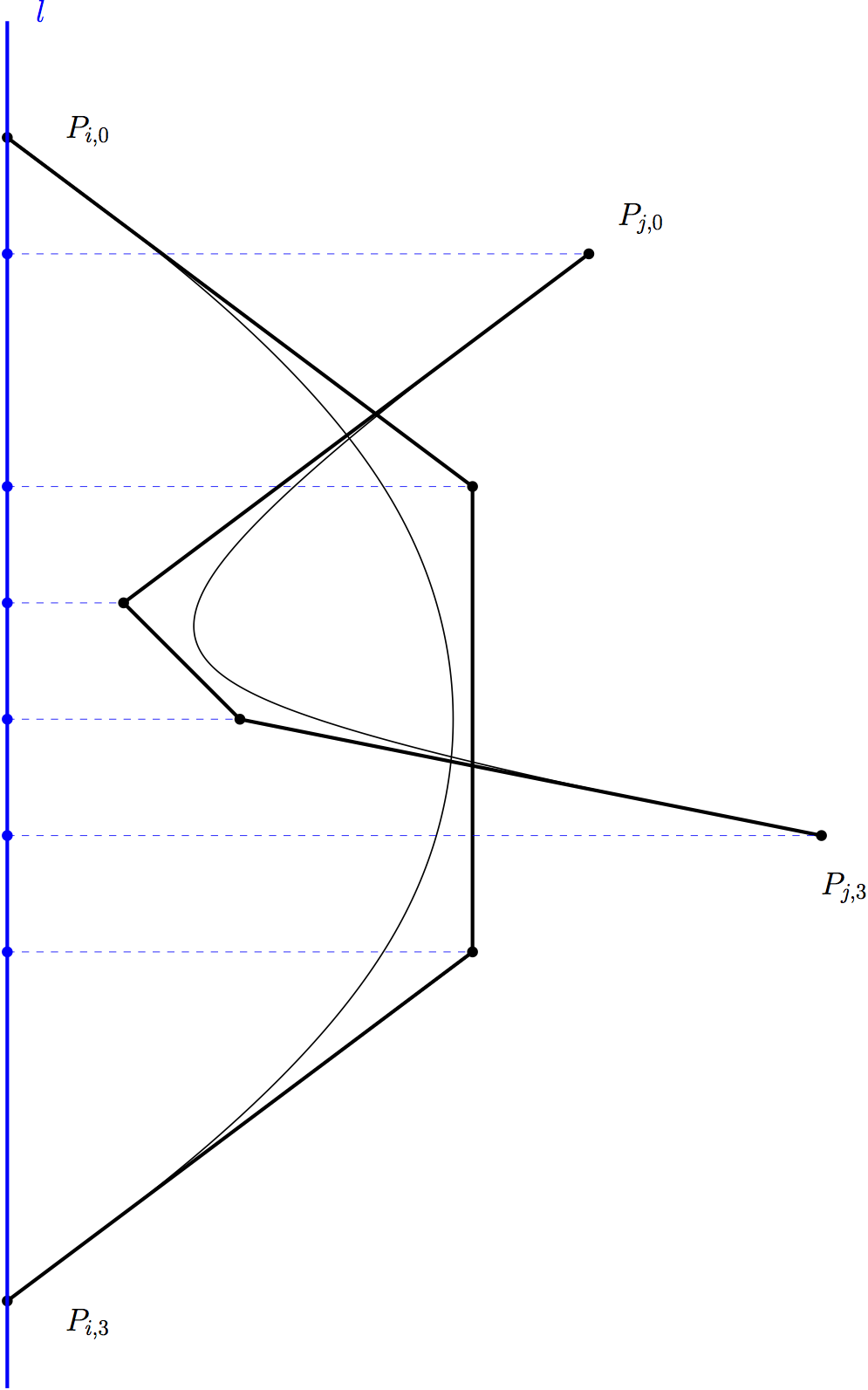}
\centering{
\caption{Generic position of intersecting patches}
}
\label{Fig.flipnew1}
\end{figure}

If $\Gamma_i(t)=B([P_{i,0},\ldots,P_{i,3}],t)$ and $\Gamma_j(t)=B([P_{j,0},\ldots,P_{j,3}],t)$ are two patches such that their control polygons have an intersection, then $\gamma_i$ joins $P_{i,0}$ to $P_{i,3}$ and $\gamma_j$ joins $P_{j,0}$ to $P_{j,3}$. Let $l$ be the line joining  $P_{i,0}$ to $P_{i,3}$, we denote $\pi_l$ the orthogonal projection on $l$. We will "sweep" along $l$. Then, $\pi_l(P_{j,0}) = \pi_l(P_{j,3})$ or not. The general case is when the two projections are distinct, i.e. $\pi_l(P_{j,0}) \neq \pi(P_{j,3})$ (see fig. \ref{Fig.flipnew1}). 

By considering an intersection test based only on interval sorting, we speed up the contour self-intersection detection. Note that a full interval sorting is achieved only at the initialization of the active contour patches and when a flip procedure is operated. During active contour evolution, the patches ordering varies insignificantly, thus a local sorting by insertion on neighboring patches is locally performed, which drastically reduces the computational cost.

The following proposition bounds the cost of the flip procedure:
\begin{lemma} \label{cost4}
The dominant computational cost of the flip procedure is related to the sorting of the FF-active contour patches. 
The cost of the sorting step is bounded by $\mathcal{O}(N\log{N})$ comparisons on average where  $N$  is the current total number of the active contour patches.
\end{lemma}
Note that the contour patches sorting is achieved once at the initialization of the active contour patches and after each flip to reorganize the patches of the disconnected contour components.

\begin{figure}[h]
\centering
\subfigure[]{\includegraphics[width=2.8cm]{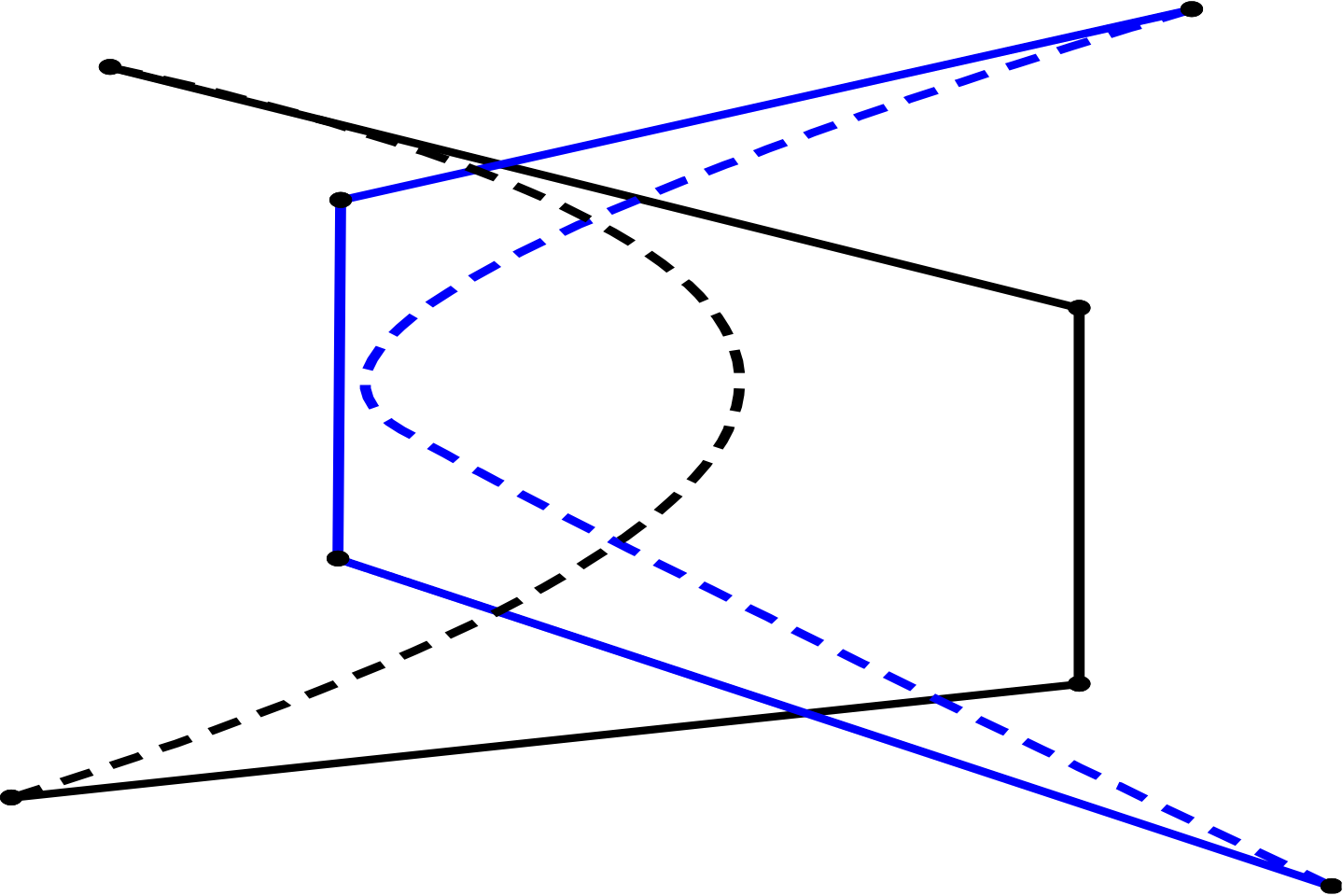}}
\subfigure[]{\includegraphics[width=2.8cm]{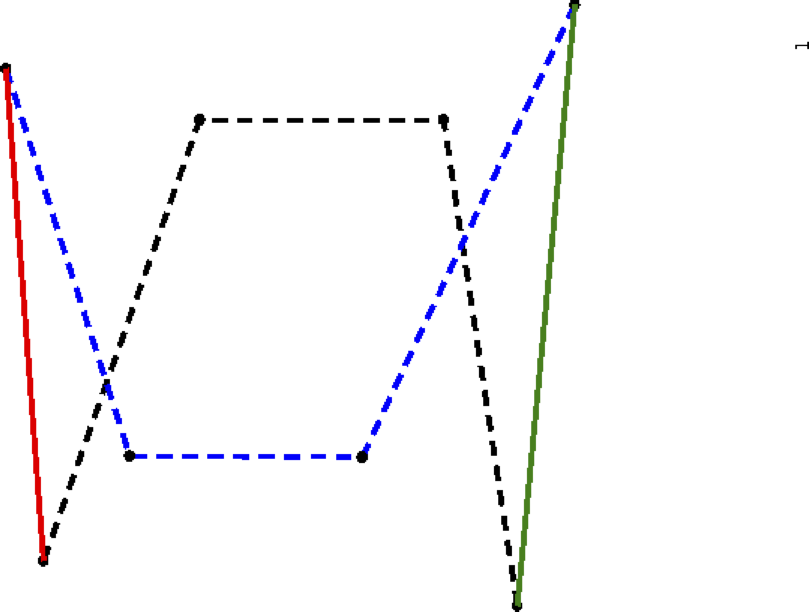}}
\caption{Flip transformation. (a) Intersection of two patches; (b) Flipping of their control points; line segments are B\'ezier curves}
\label{Fig.flip}
\end{figure}

When convex hulls intersection is detected, the control points of the corresponding patches $(\Gamma_i,\Gamma_j; i\!<\!j)$ are flipped (see Fig.~\ref{Fig.flip}). The flip procedure is performed locally by considering the endpoints control points of the intersecting patches and their immediate neighbors. The new resulting patches are formed by reordering the control points (see Algorithm~\ref{algo.flip}); which results in disconnecting the original Free Form contour into two connected components. The patches thereof are resorted so as to preserve the inherited connectivity, in a way that the bound of lemma \ref{cost4} is probably not sharp (using the insert sorting instead of a global one on the involved patches).

This approach leads to locally separate the active contour into two connected components. Topology transformations are thus handled automatically and consistently during all the active contour evolution process.


\begin{algorithm}
\caption{Flip procedure.}
\label{algo.flip}
\begin{footnotesize}
\begin{algorithmic}
\REQUIRE FF-based active contour $\Gamma_{d,N}$, index list $L_{\Gamma_{d,N}}$ of sorted patches
\ENSURE New FF composed of two connected FFs, new index list $L_{\Gamma_{d,N'}}$
\smallskip
\STATE \textbf{Do} local test intersection on $L_{\Gamma_{d,N}}$ neighboring patches;
\IF{Non empty $\Gamma_i, \Gamma_j$ $(i<j)$ patches intersection}
\STATE Disconnect curves: $\left\{\!\!\!\begin{array}{l}\Gamma_i=B([P_{0,i},P_{1,i},...,P_{d,i}])\\  \Gamma_j=B([P_{0,j},P_{1,j},...,P_{d,j}]) \end{array}\right.$
\STATE Flip control points;
\STATE Reorder control points and create new patches:
\STATE $\left\{\!\!\!\begin{array}{l}\Gamma'_i=B([P_{0,i},P_{1,i},..,P_{k,i},P_{k+1,j},...,P_{d,j}])\\ \Gamma'_j=B([P_{k+1,i},..,P_{d,i},P_{0,j},...,P_{k,j}])$; $(k\!=\!floor(d/2))\end{array}\right.$
\STATE Insert the new patches: $\Gamma_{out}\supseteq\Gamma'_i$, $\Gamma_{inn}\supseteq\Gamma'_j$;
\STATE Resort $\Gamma_{out}$ patches by increasing $x_{i,min}$;
\ENDIF
\STATE $\Gamma_{d,N'}\leftarrow\Gamma_{out}$; ($N'<N$)
\STATE \textbf{return} $\Gamma_{d,N'},\Gamma_{inn}$ and $L_{\Gamma_{d,N'}}$;
\end{algorithmic}
\end{footnotesize}
\end{algorithm}

\subsection{Global overview of the method}
\label{Sec:Algorithm}

The active contour deformation based on free form parametrization is summarized in Algorithm~\ref{algo.deform}. The corresponding global computational cost is bounded by Lemma~\ref{costAll}.

\begin{algorithm}
\caption{Free form active contour deformation.}
\label{algo.deform}
\begin{footnotesize}
\begin{algorithmic}
\REQUIRE FF-based active contour $\Gamma$, index list $L_{\Gamma}$ of sorted patches
\ENSURE Deformed FF-based active contour $\Gamma$ with (new) index list $L_{\Gamma}$
\smallskip
\WHILE{non convergence}
\STATE Split and insert $(\Gamma, L_{\Gamma})$ (algorithm \ref{algo.split});
\STATE Flip $(\Gamma, L_{\Gamma})$ (algorithm \ref{algo.flip});
\FOR{each patch of $\Gamma$}
\STATE Sample the patch;
\STATE Compute the deformation at each sampled point of the patch;
\STATE Deform the patch;
\ENDFOR
\ENDWHILE
\STATE \textbf{return} $(\Gamma,L_{\Gamma})$;
\end{algorithmic}
\end{footnotesize}
\end{algorithm}

\begin{lemma} \label{costAll}
	The overall computational cost of the method is dominated by the flip procedure when invoked.  In case of images with topological changes, the cost is bounded by $\mathcal{O}(cN\log{N})$ with $c$ the number of the free form contour components and $N$ the total number of the active contour patches. If  the deformation of the active contour does not induce any topology changes of the free forms, then the cost is linear $\mathcal{O}(N)$ in the number of contour patches (up to the initial sorting).
\end{lemma}


\section{Free form deformation and segmentation results} \label{Sec.results}
In this section, we first consider segmentation results on toy images in order to illustrate the behavior of the method, then we present the results for free space segmentation in unknown indoor and outdoor environments.

\subsection{First results on synthesis images}

\begin{figure}[h]
\centering
\includegraphics[width=8.6cm]{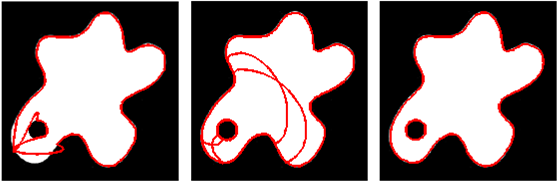}
\caption{
Free form active contour results on toy data to illustrate the effects of the different steps of the proposed algorithm. (\emph{left:}) Free form contour result, from a large number of initial patches ($N=80$), without geometric refinement, nor topology changes. (\emph{middle:}) The effect of geometric refinement (without the ability to change curve topology), starting from a small number of patches ($N=8$).  (\emph{right:}) The effect of topology changes in addition to geometric refinement. }
\label{Fig.toysdata}
\end{figure}

Figure~\ref{Fig.toysdata} illustrates, using binary shapes, the effects of the different steps of the proposed algorithm. The initial active contour is composed of $N$ linked patches located in the homogeneous area (inside the shape). One can see (left side of Fig.~\ref{Fig.toysdata}) that the final contour does not fit properly to the form, even starting from a large number of initial patches. The geometric refinement (middle side of Fig.~\ref{Fig.toysdata}) correctly overcomes this problem but does not adapt to topology changes (it loops indefinitely around the holes in the topology). The result (right side of Fig.~\ref{Fig.toysdata}) shows the adaptation of the contour to the shape when geometric refinement and topology changes are handled, enabling the contour to move into boundary concavities. This can be a first step in an object recognition procedure. 

\subsection{Free space segmentation results}

We are here interested to extract free space encompassing the robot in unknown environments while using omnidirectional visual information only. The robot visual perception is centered in the robot camera frame and is used to initiate the active contour in the close and safe neighborhood around the robot.  


Using Algorithm~\ref{algo.deform}, the global free form contour stops its evolution when almost all free form points have reached their steady state. The B{\'e}zier smoothness and the contour closure allow to well construct subjective contours and free space terminations. This appealing aspect of active contours was already revealed in early work of \cite{Kass88} showing that the same active contour that creates subjective contours can very effectively find more traditional edges in natural images.


Several experiments led in various indoor and outdoor environments, evaluate the efficiency of the proposed method to perform detection of unknown free spaces. 
The test images used in the experiments are real world images of resolution $800\!\!\times\!\!600$ acquired by a mobile robot in guided exploration.  Recall that no metric nor topology information about the robot environment is available to help the extraction process.

\begin{figure}[htbp]
\centering
\subfigure{ \includegraphics[width=4cm]{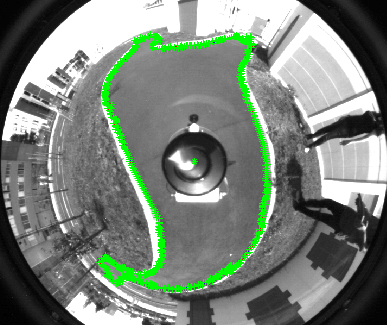}}\hspace*{0.01cm}
\subfigure{\includegraphics[width=4cm]{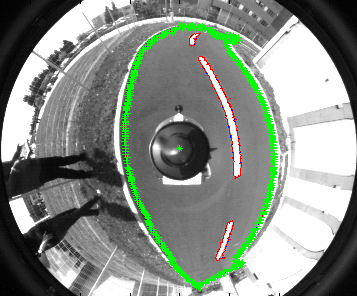}}\\
\subfigure{\includegraphics[width=4cm]{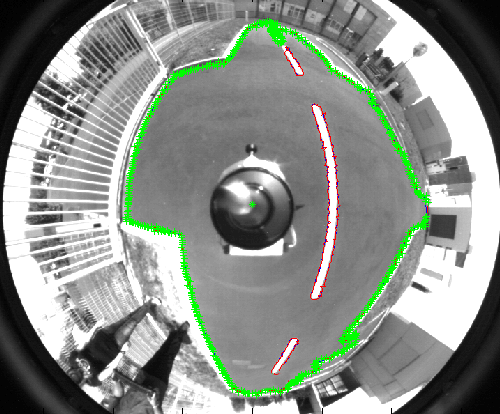}}  \hspace*{0.01cm}  
\subfigure{\includegraphics[width=4cm]{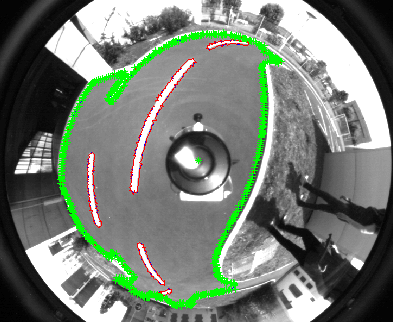}}
\caption{Omnidirectional free space extraction results obtained in unknown outdoor environments with the proposed algorithm. Connected and isolated components in the free space are displayed in red color. The initial contour is located on a circle centered at the center of the image and covering the visual robot-sensor projection in the image.}
\label{Fig.result1}
\end{figure}

Figure~\ref{Fig.result1} shows results obtained in outdoor urban environments. They give efficiency evidence of the proposed method to fit contour boundaries to the limits of the unknown free space. The initially defined free forms dynamically propagate to meet free space terminations while encompassing and separating connected components in the free space. Segmentation results obviously depend on the input edge forces which define the edge map for the active contour evolution.

\subsection{False obstacles detection}

Natural images present too many spurious edges including ground textural patterns or artificial ones. Linking edges belonging to the same object is a difficult problem, but here we make use of the free form structural topology to detect false free space terminations. The free form model is able to handle complex topology changes and is concise in describing the boundary shape of the separated components. We add a structural test on interest points if detected to classify isolated connected components as true or false obstacles. If a separated component is labelled as false obstacle, then it is merged with the main free form space. The latter extends and the robot can navigate over passable connected space areas.

Though the proposed free form approach provides rather appropriate and convincing segmentations from one single image, the use of a pairwise of two successive images is needed for the structural topology test. The latter is based on: $(i)$ Find and match interest points in the inner domain of connected components in two successive images using Harris detector; $(ii)$ the altitude of matched interest points is reconstructed by triangulation assuming the catadioptric system to be calibrated; and $(iii)$ if at least one interest matched point altitude is above a defined threshold (see test formulation~\eqref{Eq.test}), the corresponding connected component is considered as 'true positive' free space termination, else it is merged with the main free from space. The free space segmentation algorithm with false obstacles detection is shown in Algorithm~\ref{algo.main}.
\begin{equation}
\left\{\!\!\begin{array}{l}
\hbox{if  } \exists\, \mathbf{m}\!\in \Omega_{\Gamma_{inn}} \hbox{s.t. } Alt(\mathbf{m})\geq \varepsilon, \hbox{then  } \Omega_{\Gamma_{out}}\!\!:=\Omega_{\Gamma_{out}}\!\! \setminus\!\Omega_{\Gamma_{inn}}\\
\hbox{else  }  \Omega_{\Gamma_{out}}\supseteq \Omega_{\Gamma_{inn}}
\end{array}\right.
\label{Eq.test}
\end{equation}
where ${\displaystyle \Omega_{\Gamma_{inn}}}$ (resp. ${\displaystyle \Omega_{\Gamma_{out}}}$) is the compact domain delineated by the inner connected component $\Gamma_{inn}$ isolated by flipping contour patches (resp. by the main outer component $\Gamma_{out}$ of the active contour). $\mathbf{m}$ is a matched interest point located in the inner domain ${\displaystyle \Omega_{\Gamma_{inn}}}$, $Alt$ is a triangulation function computing the 3D point altitude and $\varepsilon$ is a defined altitude threshold including computation uncertainty and depending on the robot ability to move over obstacles.


\begin{algorithm}
\caption{Free form based free space segmentation.}
\label{algo.main}
\begin{footnotesize}
\begin{algorithmic}
\REQUIRE Omnidirectional image $I_k$, FF segmented free space of $I_{k-1}$
\ENSURE Free space $\Omega_{FS}$ of $I_k$
\smallskip
\STATE 1. Compute the edge map of $I_k$ using a Canny detector;
\STATE 2. Free form initialization;
\STATE 3. Free form deformation (Algorithm~\ref{algo.deform}): 
\STATE \quad $\Gamma(I_k):=\Gamma_{out}\oplus\sum_j \Gamma_{inn,j}$;
\STATE \quad ${\displaystyle \Omega_{FS}:=\Omega_{\Gamma_{out}}\!\!\setminus\cup\{\Omega_{\Gamma_{inn,j}}\}}$;
($\Omega_{\Gamma}$ is the inner domain of FF $\Gamma$);
\IF{Non single component free form $\Gamma(I_k)$}
\FORALL{$j$ \textbf{s.t.} $\Gamma_{inn,j}\subset \Gamma(I_k)$}
\STATE Compute Harris interest points in  $\Omega_{\Gamma_{inn,j}}$;
\STATE Search for inner matched interest pts in $\Omega_{\Gamma(I_{k-1})},\Omega_{\Gamma(I_k)}$;
\IF{Non empty matched interest points}
\STATE Triangulate the altitude of the matched keypoints;
\IF{altitude zero of all inner matched points (Eq.~\ref{Eq.test})}
\STATE Merge $\Omega_{\Gamma_{inn,j}}$ in the free space domain; 
\STATE $\Omega_{FS} := (\Omega_{FS}\supseteq\Omega_{\Gamma_{inn,j}})$;
\ENDIF
\ENDIF
\ENDFOR
\ENDIF
\STATE \textbf{return} the free space $\Omega_{FS}$;
\end{algorithmic}
\end{footnotesize}
\end{algorithm}

\begin{figure}[htbp]
\centering
\includegraphics[width=9cm]{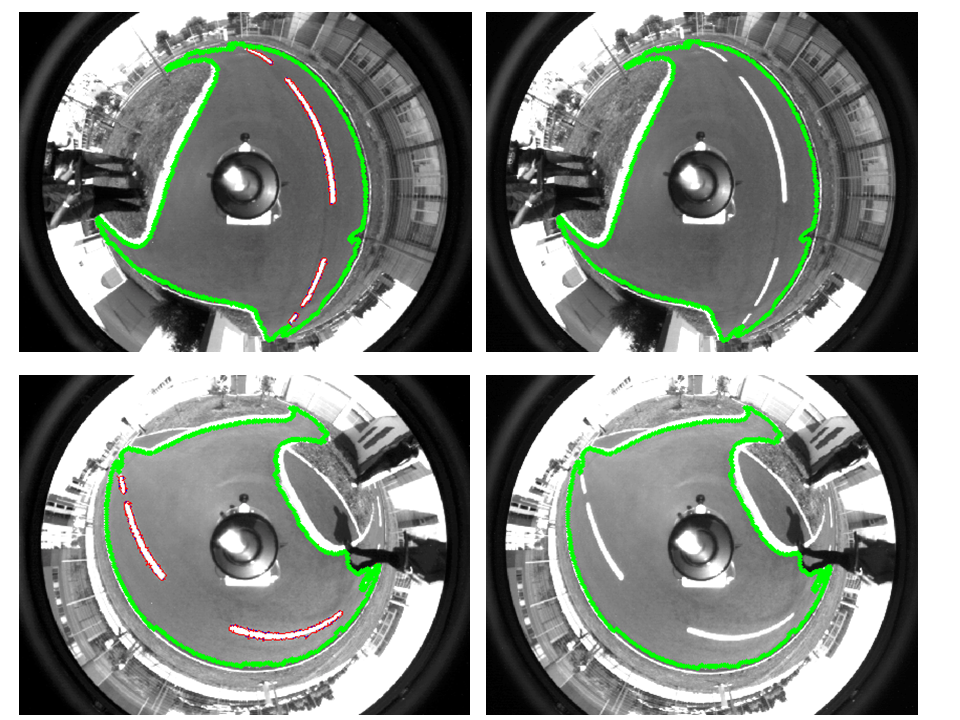}
\caption{Free space segmentation and false obstacles detection in outdoor urban environments. In the left column, connected components are enclosed by the free form algorithm. The right column shows the results after the structural topology test where false obstacles are removed.}
\label{Fig.obst1}
\end{figure}

\begin{figure}[htbp]
\centering
\includegraphics[width=8.7cm]{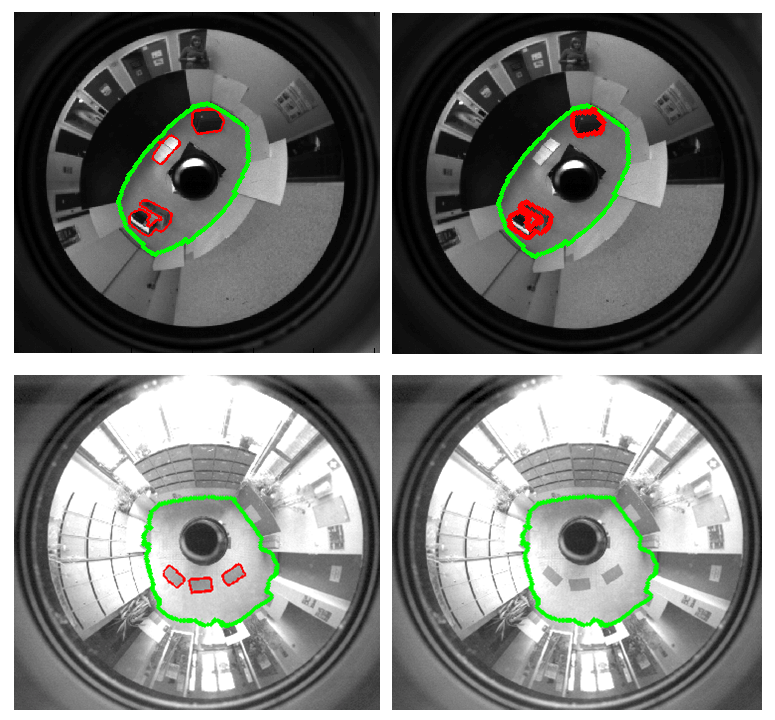}
\caption{Free space segmentation and false obstacles detection in indoor environments. In the left column, connected components are enclosed by the free form algorithm. The right column shows the results after the structural topology test where false obstacles are removed.}
\label{Fig.obst2}
\end{figure}

Figures~\ref{Fig.obst1} and \ref{Fig.obst2} depict free space segmentation results in indoor and urban outdoor environments, where several connected components of the free form active contour are enclosing white road lines or objects with varying altitudes located on the ground. The structural test allows to eliminate false obstacle candidates in the robot surroundings and thus extends its free space perception to relevant free space terminations.

\section{Comparative results and discussion}

To validate the interest of the free form method within the active contours framework, we discuss in the following its performance in comparison with traditional active contour approaches. The latter are based on energy minimization \cite{Kass88, Williams92, Xu98, Sum07, Wu10} or on resolution of an implicit variational scheme \cite{Bogdanova07, Sethian99}. Parametric active contours that by minimization deform on the domain according to the influence of internal and external forces, are very efficient with regard to processing time but are not (or rarely) able to handle topology changes. Geometric active contours that express optimal contour deformation as a solution of a partial differential equation, are topologically flexible, but are not time efficient though some numerical techniques have been developed \cite{Sethian96}  to make the level set numerically robust and efficient.

Work \cite{Merveilleux11b} adapted and compared the energy minimizing and level set methods for finding adequate approximations of visual free space in images.  We extend this comparative work to free from representation of active contours and will show that it combines advantages of both parametric and geometric methods.

\subsection{Brief description of state-of-the-art methods}

A parametric active contour is a curve defined parametrically by $\Gamma(s)$ and moved such that it minimizes an energy functional \eqref{Eq.parm}:
\begin{equation}
{\displaystyle E^{\star}_{\Gamma} = \min_{\Gamma} \left\{ \int_{\Gamma}\! E_{int}(\Gamma(s)) + E_{ext}(\Gamma(s))\, ds \right\}}
\label{Eq.parm}
\end{equation}
where $E_{int}$ and $E_{ext}$ denote respectively the internal energy of regularization and the external energy derived from the image. Different energy formulations and the underlying force fields driving the active contour have been proposed. Examples include balloon and pressure force model \cite{Cohen91}, distance transform force model \cite{Cohen93} and gradient vector flow model \cite{Xu98}. 

\begin{figure}[htbp]
\centering
\subfigure[]{\includegraphics[width=.2\textwidth]{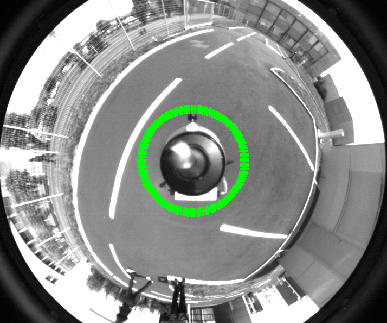}}\hfil
\subfigure[]{\includegraphics[width=.2\textwidth]{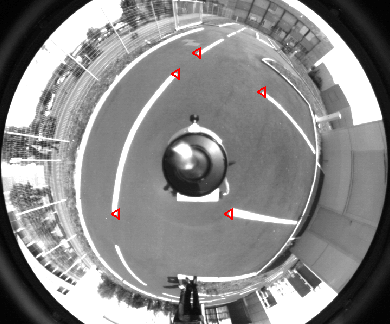}}\hfil
\subfigure[]{\includegraphics[width=.2\textwidth]{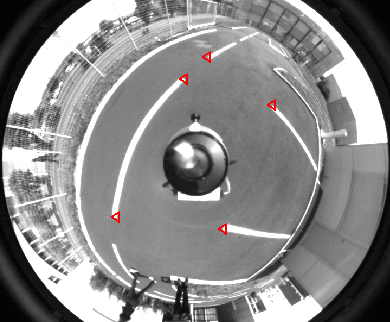}}\hfil
\subfigure[]{\includegraphics[width=.2\textwidth]{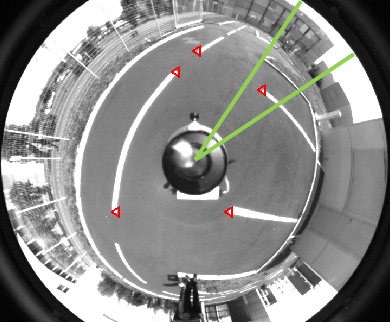}}
\caption{(a) Active contour initialization;  (b,c) two successive images acquired by a robot, and (d) matched interest points located in the radial zone enclosing an active contour point.}\label{Fig.triangul}
\end{figure}

The active contour model for free space segmentation developed in \cite{Merveilleux11a} is an iterative parametric method able to handle a real-time omnidirectional free space approximation where the problem of falsely classified obstacles is solved thanks to a new functional energy formulation. The latter considers a local test based on the intuitive idea that the observation by a robot of two successive free space terminations in the same direction is congruous only if the first observed termination belongs to false obstacles. The external energy (Eq.~\eqref{Eq.Eim}) includes an altitude function of matched interest points (using Harris detector) in the vicinity of the moving interface and superposes their altitude to the edge map generated by Canny edge detector. If two keypoints of zero altitude are successively located in a small radial zone normal to the contour propagation (see Fig.~\ref{Fig.triangul}-(d)), the contour continues its evolution over the first edge considered as a virtual obstacle.
\begin{equation}
E_{im}(\mathbf{v}) \!=\!\! \left\{ \!\!\!\!\begin{array}{l}
\left | \nabla I(\mathbf{v})\right | \left(-1+sign(\epsilon-Alt(\mathbf{m}))\right) \hbox{ if } \exists\ \mathbf{m} \in R_{\mathbf{v}}  \\
\quad \\
\left | \nabla I(\mathbf{v})\right |  \,\, \hbox{ else,} 
\end{array}
\right.
\label{Eq.Eim}
\end{equation}
where $\mathbf{m}$ is a matched interest point detected in $R_{\mathbf{v}}$, a radial zone containing the active contour point $\mathbf{v}\in\Gamma(I)$. $Alt$ is the altitude function with $\varepsilon$ threshold of obstacle-passing. 
The proposed model also includes dynamic geometric refinement based on a cost distance function that allows the contour to deform and adapt to large or more complex shapes.

Geometric active contours are alternative models to parametric contours, based on the 
fundamental concept of level set methods. The latter specify the active contour in an implicit form function which evolution is modeled by the Hamilton-Jacobi nonlinear equation \eqref{Eq.HJ}. 
\begin{equation}
\left\{\!\!\!
\begin{array}{l}
\psi_t(\mathbf{x},t)+c(\mathbf{x})|\nabla\psi(\mathbf{x},t)|=0,\, \, \hbox{for }\, \mathbf{x}\in\!\Omega, t\in\!(0,+\infty) \\
\quad\\
\psi(\mathbf{x},0) = \psi_0(\mathbf{x}),\, \,\, \hbox{ for }\, \mathbf{x}\in\!\partial\Omega
\end{array}\right.
\label{Eq.HJ}
\end{equation}
where $\psi(\mathbf{x},t)$ is the implicit scalar function on domain $\Omega$ with initial condition $\psi_0$ representing the initial front $\Gamma(t\!\!=\!\!0)$. The normal speed $c(\mathbf{x}):\Omega\rightarrow\mathbb{R}$ of the moving interface is determined from the shape gradients.

Approximate numerical solutions to the stationary variational scheme (where $c$ is monotone in \eqref{Eq.HJ}) have been efficiently produced using fast marching techniques \cite{Sethian96, Cristiani07}. The latter concentrate the scheme resolution only in a small neighborhood of the front to avoid useless
computations. This is done by dividing the discrete domain into three subsets: far nodes, accepted nodes, and narrow band nodes. 
The narrowband nodes are the nodes lying in the neighborhood of the deforming contour and where a solution is actually computed. The accepted nodes are those already traversed by the front and where an approximate solution was previously evaluated (their values cannot change). The far nodes are the remaining nodes not yet reached by the propagating front and where no approximate solution was ever estimated.

Work \cite{Merveilleux11b} added a new category of nodes, labelled \emph{potentially on contour} nodes, which are located in the narrow band and its immediate neighboring with the following property: When a narrow band point is accepted, if its neighboring points are on edges (i.e. $c$ is close to zero in these points), they are considered as potentially on contour nodes and the front stops its propagation at these points.

\subsection{Experimental results analysis}

The experimental protocol consists in guiding a mobile robot to explore indoor and outdoor urban environments, and to extract the visual free space using the active contour models presented in the previous section and the free form model proposed in this paper. The results are assessed with respect to the final shape of the free space terminations and time convergence. 

\begin{figure}[h]
\centering
\includegraphics[width=8.7cm]{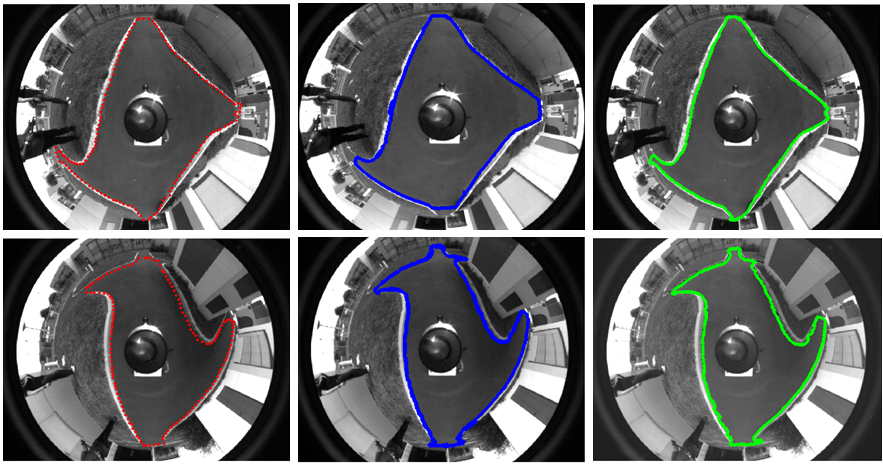}
\caption{Comparative active contour results on omnidirectional images from robot-collected dataset in outdoor environments (where the free space is a single connected component), 
using (\emph{left}) parametric model, (\emph{middle}) level-set method and (\emph{right}) free form model.}
\label{Fig.comp1}
\end{figure}

Figure~\ref{Fig.comp1} shows quite similar results in free spaces of simple shapes with an improvement of the final shape smoothness in the level set and free form models due to the intrinsic regularity of, respectively, the level set solution and B{\'e}zier parametrization.  Note that the regularization weighting parameters of the internal energies have been manually tuned and optimized at the initial state on a set of test images. The latter do not seem, as we could expect, sufficient to maintain curve smoothness in spaces of arbitrary shapes.

\begin{figure}[h]
\centering
\includegraphics[width=8.6cm]{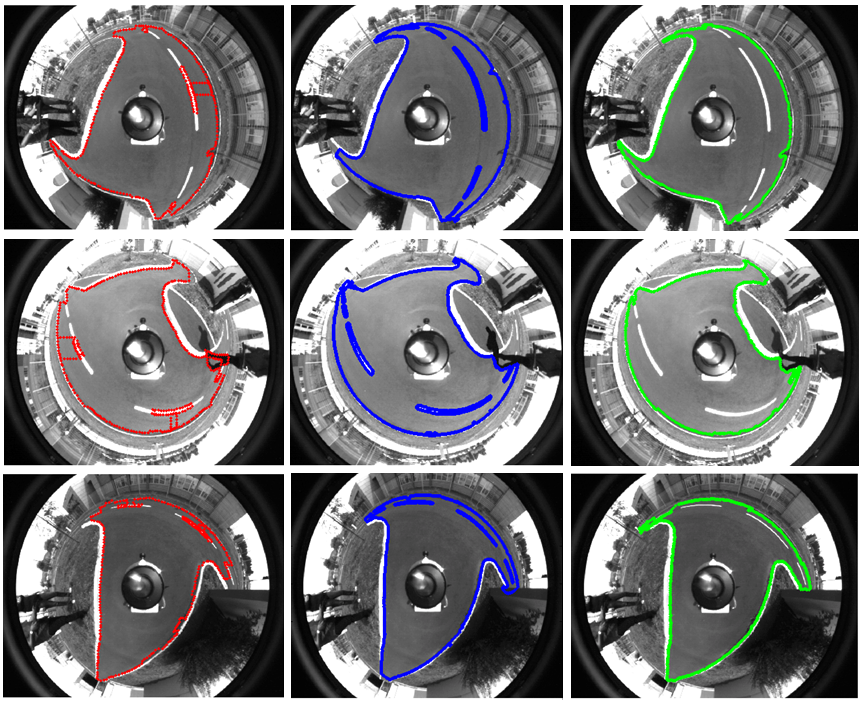}

\caption{Comparative active contour results on omnidirectional images from robot-collected dataset in outdoor urban environments (where the free space includes multiple connected components), using (\emph{left}) parametric model, (\emph{middle}) level-set method and (\emph{right}) free form model. Note situations such \emph{(top and mid-left)} where the parametric active contours do not completely overpass the road line delimitation. This is due to the spacing between the interest points (located at the ends) of the line, which is too large to include these points in a cone of detection around the contour points.}
\label{Fig.comp2}
\end{figure}

\begin{figure}[h]
\centering
\includegraphics[width=8.4cm]{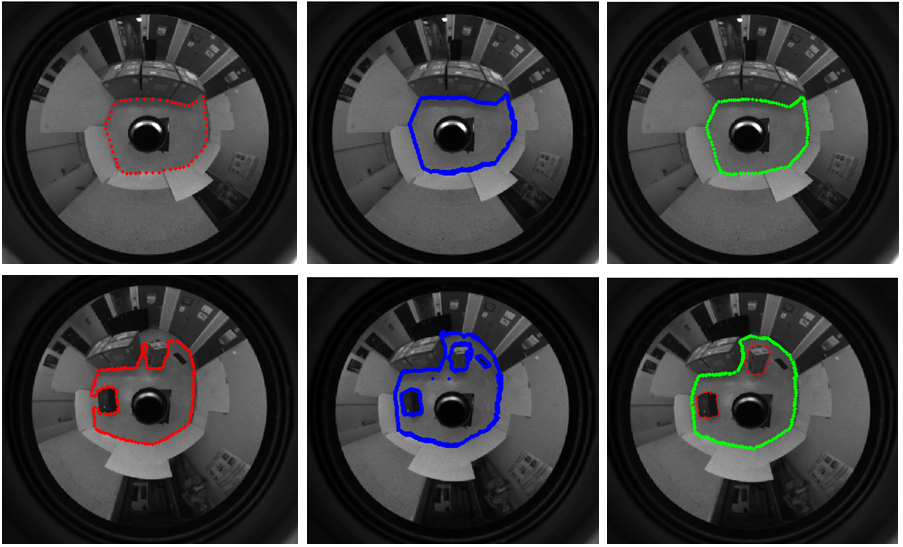}
\caption{Comparative active contour results on omnidirectional images from robot-collected dataset in indoor cluttered environments, using (\emph{left}) parametric model, (\emph{middle}) level set method and (\emph{right}) free form model. Objects of various heights (sheets, boxes) were deliberately introduced in the space of navigation of the robot, to test the behavior of the assessed models. By using the structural topology test, the robot extents its perception of the free space by removal of the isolated free form components that encompass passable objects.}
\label{Fig.comp3}
\end{figure}

In more complex environments (Fig.~\ref{Fig.comp2} and \ref{Fig.comp3}) yielding topology changes, the free form and level set models highly outperform the parametric one. An intrinsic behavior of geometric deformable models is the ability to handle topology changes. The latter are also automatically considered in the free from model including the intersection test. Though the optimized parametric model \cite{Merveilleux11a}  integrates local false obstacle detections, a single closed contour can hardly manage situations with large spurious free space terminations, as illustrated in Fig.~\ref{Fig.comp2}.  The behavior of the parametric model is most dependent on the presence and distribution of interest points near the contour. A lack of interest points extracted prevents disambiguate the entire junction.

Note however that no semantics of objects is present for the robot. So if the objects are too close to walls so that the robot can not see behind the object, they can not be isolated and enclosed by the free form method. They will be considered as free space termination regardless of their altitude.

Finally, we consider the computational cost of the proposed free form method and compare it with that of the parametric and level set methods. Table~\ref{Tab.comp} shows comparative results with respect to the processing time of the algorithms. The active contours were implemented using \emph{Matlab} and \emph{\sc c/c++} for real-time robot perception and navigation. Each method parameters were optimized to give optimal performance of the final segmented shape of the free space. Our experiments indicate that (a) the representation of the final shape of the free space is better with level set and free space models, and (b) the time elapsed is almost similar for the parametric and free form models whereas it increases significantly with the level set model. This processing time of the free form model improves in environments of simple shapes with time gain of order 2 (respectively 4) relatively to parametric (respectively level set) model. 
The free form algorithm is almost linear with respect to the number of patches. The main computational cost of the free form method is to sort the patches of the contour (which cost is $N\log N$ with $N$ the total number of contour patches) but this is done only at the initialization and after each topology changes.

\noindent
\begin{table}[h]
\caption{\label{Tab.comp} Compared computational cost between the implemented (using \emph{\sc c++}) active contour models.}
\centering
\begin{small}
\begin{tabular}{|p{1.4cm}|p{.8cm}|p{.9cm}|p{.9cm}|p{1.1cm}|p{1.1cm}|}
\hline
\footnotesize{Input} & \hspace{-.1cm}\footnotesize{Method} & \footnotesize{Initial} & \footnotesize{Final} & \footnotesize{Final no} & \hspace{-.1cm}\footnotesize{Time \emph{(ms)}} \\
\footnotesize{image} &  & \multicolumn{2}{c|}{\footnotesize{\mbox{num. of cont. pts}}} & \footnotesize{\mbox{of iter.}} & \footnotesize{\mbox{converg.}}\\\hline     
      \hspace{-.15cm}\multirow{3}{*}{\includegraphics[width=1.5cm,height=1.1cm]{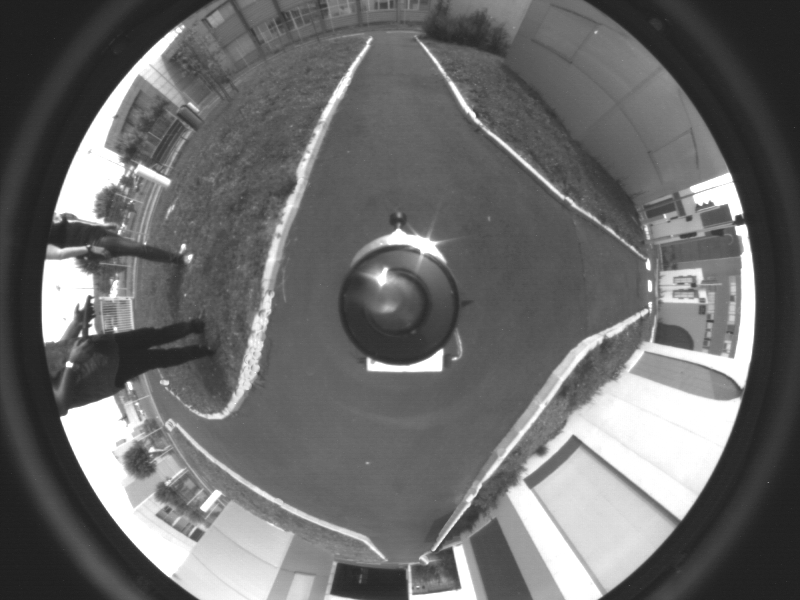}} & \textbf{PM} & 40 & 143 & 159 & 45 \\ \cline{2-6}
      & \textbf{LS} & 516 & 2077 & 98320 & 95 \\  \cline{2-6}
      & \textbf{FF} & 40 & 432 & 229 & \textbf{22} \\ \hline
      \hspace{-.15cm}\multirow{3}{*}{\includegraphics[width=1.5cm,height=1.09cm]{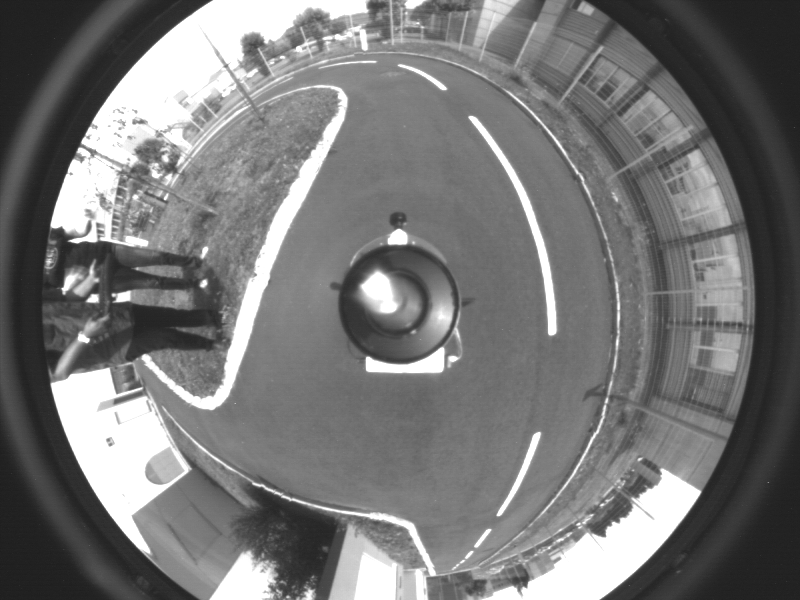}} & \textbf{PM} & 40 & 312 & 186 & \textbf{42} \\ \cline{2-6}
      & \textbf{LS} & 516 & 3528 & 114579 & 168 \\  \cline{2-6}
      & \textbf{FF} & 40 & 852 & 252 & 48 \\ \hline
\end{tabular}
\smallskip
\caption*{Typical original input images of simple (\emph{top}) and complex (\emph{bottom}) environments are shown on the left. PM, LS and FF stand for respectively parametric, level set and free form models of the active contour. Note that \emph{i)} The level set method requires a large number of initial points for the validity of the finite difference scheme approximation; and \emph{ii)} it moves one point at each iteration, while the other two methods move the entire contour at each iteration.}
\end{small}
\end{table}

\section{Conclusion} \label{Sec.conclu}

This paper presented a fast technique for active contours based on free form modeling and deformation. It was applied to visual free space segmentation using omnidirectional imagery from a catadioptric monocular system with the aim of achieving robot autonomous exploration and navigation in unknown environments. The proposed method can easily adapt to complex free space shapes that can form holes, split to form multiple boundaries, or merge with other boundaries to form a single surface. It incurs local control of deformation that is very fast and smooth deformations of arbitrary shapes. There is no need to re-parameterize the model as it undergoes significant changes in shape.

The proposed free form method was also compared to recently developed real-time deformable models, based on parametric and level set formulations. The resulting free form algorithm requires significantly lower computational cost, that is advantageous over the original methods for real-time image processing and robot navigation.

The Free Form contour deformation has been applied in this work to the perception of free space surrounding the robot, but obviously it can also solve segmentation problems in general.  Experimental results with robot exploring unknown indoor and urban outdoor environments show the efficiency of the method to perform, in real-time, rather convincing approximations of the robot free surroundings, even if the latter is highly dynamic or cluttered by false obstacles. Our ongoing work is interested in the skeletization of the segmented free space for topological mapping and autonomous robot exploration.





\end{document}